# Level Playing Field for Million Scale Face Recognition

Aaron Nech    Ira Kemelmacher-Shlizerman

Paul G. Allen School of Computer Science and Engineering

University of Washington

{necha, kemelmi}@cs.washington.edu

## Abstract

*Face recognition has the perception of a solved problem, however when tested at the million-scale exhibits dramatic variation in accuracies across the different algorithms [11]. Are the algorithms very different? Is access to good/big training data their secret weapon? Where should face recognition improve? To address those questions, we created a benchmark, MF2, that requires all algorithms to be trained on same data, and tested at the million scale. MF2 is a public large-scale set with **672K identities and 4.7M photos** created with the goal to level playing field for large scale face recognition. We contrast our results with findings from the other two large-scale benchmarks MegaFace Challenge and MS-Celebs-1M where groups were allowed to train on any private/public/big/small set. Some key discoveries: 1) algorithms, trained on MF2, were able to achieve state of the art and comparable results to algorithms trained on massive private sets, 2) some outperformed themselves once trained on MF2, 3) invariance to aging suffers from low accuracies as in MegaFace, identifying the need for larger age variations possibly within identities or adjustment of algorithms in future testings[1]*

## 1. Introduction

> All that glisters is not gold
> Often have you heard that told.
>
> — William Shakespeare

According to Google Scholar, in just the year 2016, 938 face recognition algorithms were published, 34 patents were filed, and dozens of face recognition startups were established. Given such colossal resources, let's say one wishes to create an application that uses the best face recognition algorithm out there, how would they know which algorithm is better to implement or buy?

Two prominent problems in face recognition are verification and identification. Verification is the problem of verifying if two images of faces are the same person, while identification is the problem of determining the identity of a face image. Public benchmarks help rank algorithms and indeed on smaller test sets, e.g., Labeled Faces in the Wild (LFW) [9], YouTube Faces Database (YFD) [30], and IJB-A [14], computers have reached phenomenal accuracies in both verification and identification. Recent large-scale benchmarks [11, 8], however, consistently indicate these problems are not solved on a global scale (where millions, even billions of identities are to be distinguished). Moreover, there is a dramatic difference in accuracy across the algorithms.

High variation in accuracy across algorithms raises an interesting question. Is it really that some particular algorithm outperforming others or access to big/good training data is the key to success? That's what we aim to investigate in this paper. The idea is the following: let's create a training set which can be used by everyone (publicly available), require groups to train **only** on that data and evaluate at scale (unlike [11, 8] which allow to train on any data, including private sets). This should potentially level the playing field and benchmark the actual algorithms rather than the data they use. Creating a large public training set is a challenge though.

The ImageNet competition [24] showed that neural networks [15] approaches dominate, and tend to perform better as 1) deeper networks are developed and 2) more data is provided to accurately tune network weights. It is important, therefore, for a benchmark to provide big enough data for algorithms to be successful. Private companies have access to millions of labeled identities, however these can not be made public. [21, 8, 31] are among the biggest public sets (Table 1). One interesting direction would be to combine those into a single training set. Instead we chose to create a new set from Flickr photos due to: 1) most public sets (training and testing) are photos of celebrities; to remove dataset bias we chose to use mostly non-celebs for training, 2) based on the success of FaceNet [25], in [11], we aimed for large number of identities (to span the diversity in hu-man population); max number of identities before MF2 was 100K, while MF2 has 672K. Given these considerations we discuss several key aspects in the paper:

1. An *automatic* 672K identity labeling algorithm. For contrast, ImageNet and the Google Open Images [1] sets include 10K classes. Human annotation of small datasets is more accurate, however, scaling up to millions photos is challenging [19]. Using compact embeddings, we are able to cluster millions of faces provided assumptions based on the data structure hold.

2. Baselines trained on MF2 and the different approaches in dealing with 672K classes during training.

3. Benchmarking with MF2 and provide insights contrasting with MegaFace and MS-Celebs-1M challenges.

---

[1] Benchmark is updated frequently and available at http://megaface.cs.washington.edu.

| Dataset | Celebrity? | Identities | Size |
|---|---|---|---|
| LFW | Yes | 5K | 13K |
| FaceScrub | Yes | 530 | 106K |
| YFD | Yes | 1.5K | 3.4K Videos |
| CelebFaces | Yes | 10K | 202K |
| UMDFaces | Yes | 8.5K | 367K |
| CASIA-WebFace | Yes | 10K | 500K |
| MS-Celeb-1M | Yes | 100K | 10M |
| VGG-Face | Yes | 2.6K | 2.5M |
| **Ours** | **No** | **672K** | **4.7M** |
| Facebook† | No | 4K | 4.4M |
| Google † | No | 8M | 200M+ |
| Adience | No | 2.2K | 26K |

Table 1. Representative face datasets that can be used for training. † Denotes private dataset.

## 2. Related Work

We focus on training datasets and benchmarks, labeling of large scale data, and large-scale training.

### 2.1. Data and Benchmarks

Table 1 summarizes representative datasets for training. Notably, VGG-Face, CASIA-Webface [32], UMD-Faces [2], or MS-celebs-1M [8] would be typically used for training (since they are bigger than others), unless access to private data is available. Private sets, e.g.,Google, Facebook, or governmental databases can not be made public. Private data can get to as many as 8M identities and 200M+ photos [25], while the largest public dataset has 100K identities and 10M photos [8]. Public sets are mostly created from celebrity photographs, thus the labeling problem involves harvesting many celebrity names, collecting photographs pertaining to these labels, and utilizing a combination of automated and manual data verification and cleaning to produce the final data set. Since the label is known, well-curated websites such as IMDB, knowledge graphs such as Google Freebase, and search engines are leveraged.

By training only on celebrity photographs, we risk constructing a bias to particular photograph settings. For example, it is reasonable to assume that many celebrity photographs were obtained with high quality professional cameras, or that many celebrities photographs are not of children. Algorithms learned with this bias may perform differently when tested on photographs of non-celebrities.

The most recent benchmarks in large-scale face recognition are MegaFace [11] and MS-celebs-1M [8]. Both allow training on any data. Interestingly results from the two benchmarks are consistent, i.e., highest score is around 75%, and there is dramatic difference across algorithms. In this paper, we experiment with a fixed training set (no outside data is allowed).

### 2.2. Large-scale labeling

Labeling million-scale data manually is challenging and while useful for development of algorithms, there are almost no approaches on how to do it while controlling costs. Companies like MobileEye, Tesla, Facebook, hire thousands of human labelers, costing millions of dollars. Additionally, people make mistakes [10, 5] and get confused with face recognition tasks, resulting in a need to re-test and validate further adding to costs. We thus look to automated, or semi-automated methods to improve the purity of collected data.

There has been several approaches for automated cleaning of data. [21] used near-duplicate removal to improve data quality. [16] used age and gender consistency measures. [3] and [33] included text from news captions describing celebrity names. [20] propose data cleaning as a quadratic programming problem with constraints enforcing assumptions that noise consists of a relatively small portion of the collected data, gender uniformity, identities consist of a majority of the same person, and a single photo cannot have two of the same person in it. All those methods proved to be important for data cleaning given rough initial labels, e.g., the celebrity name. In our case, rough labels are not given. We do observe that face recognizers perform well at a small scale and leverage embeddings [29] to provide a measure of similarity to further be used for labeling.

### 2.3. Large-scale training

Large-scale training optimization considers large numbers of samples per class where batching and online approaches, e.g., stochastic gradient descent, are valuable [4]. [18] suggested to augment the number of samples per identity using domain specific techniques, such as expression altering or pose warping. In such case, samples can be parametrized and trained more effectively.

A complementary problem is how to scale and optimize training in case the number of classes (rather than samples in each class) is big. Fully connected softmax networks [21], performed training on only 2.6K classes (identities) at once. Optimizing directly for embeddings, without predicting all classes at once, showed to scale better [25]. MS-celebs-1M [8] leveraged smaller pre-trained networks as starting points for larger data, i.e., bootstrap a network trained on 2.6K identities to train further on 10K+ identities.

## 3. Data Collection for MF2

To create a data set that includes hundreds of thousands of identities we utilize the massive collection of Creative Commons photographs released by Flickr [26]. This set contains roughly 100M photos and over 550K individual Flickr accounts. Not all photographs in the data set contain faces. Following the MegaFace challenge [11] and [6], we sift through this massive collection and extract faces detected using DLIB's face detector [12]. To optimize hard drive space for millions of faces, we only saved the crop plus 2 % of the cropped area for further processing. After collecting and cleaning our final data set, we re-download the final faces at a higher crop ratio (70%). As the Flickr data is noisy and has sparse identities (with many examples of single photos per identity, while we are targeting multiple photos per identity), we processed the full 100M Flickr set to maximize the number of identities. We therefore employed a distributed queue system, RabbitMQ [22], to distribute face detection work across 60 compute nodes which we save locally. A second collection process aggregates faces to a single machine. In order to optimize for Flickr accounts with a higher possibility of having multiple faces of the same identity, we ignore all accounts with less than 30 photos. In total we obtained 40M unlabeled faces across 130,154 distinct Flickr accounts (representing all accounts with more than 30 face photos). The crops of photos take over 1TB of storage. As the photos are taken with different camera settings, photos range in size from low resolution (90x90px) to high resolution (800x800+px). In total the distributed process of collecting and aggregating photos took 15 days.

## 4. Automatic Identity Labeling

Our next task is to cluster the unlabeled faces into identities. While faces are unlabeled we do have their respective Flickr IDs. The key idea, then, is that while face recognition is unsolved on large scale, it works well on a small scale, e.g., clustering small number of people in a single ID. For example [21] obtains over 99% accuracy on the LFW benchmark. Thus we first run a face recognizer on each Flickr ID. Each Flickr ID has on average 307 photos per account. Then we develop a clustering algorithm that can distinguish between the small number of identities in each ID. Our face recognizer utilizes the pre-trained VGG-Face model [21], and further augments the performance by training a triplet projection layer over the data set released by VGG-Face. A 1024-dimensional triplet embedding is used to extract euclidean distance comparable features over our entire set of 40M faces.

### 4.1. Clustering considerations and assumptions

Unlike celebrity data set construction, the identity of the individuals in Flickr data is unknown thus we must directly cluster a large set of unlabeled data instead of growing existing clusters based on search terms. The task of clustering all 40 million faces in the same space is enormous, and error prone. On a local Flickr account scope, however, face recognition can be expected to perform well with current available algorithms, thus we can achieve strong results by aggregating a series of smaller clustering problems, and run those in parallel. We also make cluster size assumptions in order to alleviate noise. In particular we make the assumptions:

1. Identities cannot be found across Flickr accounts.
2. Identity cannot be found more than once in a photo.
3. An identity must have $> Z$ photos for an integer Z.

The assumption that identities cannot be found across Flickr accounts is one to bootstrap the quality our data set relative to the algorithm used. Current available algorithms do not perform well on the millions-scale using clustering approaches on the entire unlabeled data corpus directly. When tried, the example clusters we inspected were simply incorrect as the problem was too large. Instead we made this assumption to reduce the scale of clustering to the hundreds, in which the triplet tuned VGG-Face features excelled at separating. In this way we gain information from the structure of the Flickr data itself. Additionally, this allowed us to run our clustering algorithm in parallel, drastically reducing computation time.

These assumptions are not valid for all cases. For example, celebrities may be found within the Flickr data, or "collage" images can be formed which contain multiple copies of one identity. However, in practice we found these assumptions hold well.

Under these conditions, it is also not known how many clusters (identities) exist within a particular Flickr account. Thus we limit ourselves to clustering algorithms which can discover the number of clusters. We have tried a number of approaches. Iterative clustering algorithms, such as cop-kmeans and kmeans with elbow detection [28, 17, 27], obtained poor results in the hypersphere embedding. DB-SCAN [7], a simple distance thresholding algorithm produced promising results for particular distance parameters in particular Flickr ID, but generalized poorly across multiple Flickr IDs due to variations in embedding distance. The layout (and therefore relative distances) of faces in the embedding varies by ID. For example, for some IDs a set of faces may exhibit an average distance of 0.90, and same-identity faces are on average 0.14 apart which is perfect separation, but these specific numbers are not observed across Flickr accounts making it hard to generalize. Using strict DBSCAN [7] or absolute distance thresholding also fails to encode our assumptions such as that multiple instances of the same identity cannot be in the same photo.

### 4.2. Our clustering algorithm

We therefore created our own modified clustering algorithm leveraging a simple relative-distance thresholding model, but also incorporating our assumptions. This algorithm can be broken down into the following steps, treating the embedding as a graph of faces, where edges signify common identity:

1. The graph of faces within a Flickr account is initialized with no edges between faces.
2. A no-link constraint matrix is constructed: $N \times N$ binary matrix, where $N$ is the number of faces in the Flickr account. It is populated with the results of our assumptions, i.e., entry at $i, j$ signifies that faces $i$ and $j$ can be linked with an edge signifying they are the same identity.
3. Compute $D$, where $D$ is the average pairwise euclidean distance in the Flickr account (across all pairs of faces), using the feature embedding.
4. For every pair of faces $i, j$ we threshold on parameter $\beta$ as follows: if their entry in the no-link matrix is false, and their distance is less than $\beta D$, place an edge between the faces.
5. Obtain connected components $C$ in this new graph.
6. For every connected component $c_i \in C$, if the size of the connected component is less than $Z$, remove it from the graph as it is too small to be an identity. In our work we choose a minimum component size $Z = 3$.
7. Save each connected component remaining as a cluster (identity).

The parameter $\beta$ was chosen automatically via validation with ground truth (on independent labeled sets LFW and FGNET), more details in Sec. 4.4.

### 4.3. Clustering Optimization:

We validate our clustering approach with ground truth (on independent labeled sets LFW and FGNET). The above algorithm produced over 85% purity, i.e., correct identities. There are two key observations and patterns, however, in the clustering embedding. Based on those we have further created two noise reduction mechanisms which increased the accuracy to 98% purity on both LFW and FGNET sets.

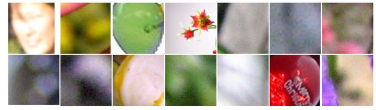

Figure 1. Example of a noise cluster containing "garbage" images (low resolution, non-faces). These tend to appear in clusters together as the embedding places them in close proximity. However, the average pairwise distance of such a cluster is much higher than a valid identity.

We identified two types of noise that survive the initial clustering algorithm:

- **Garbage clusters.** The embedding tends to place low resolution faces and non-faces (e.g. noise from the detector) in close proximity, and thus they end up forming an identity. Example in Figure 1.

- **Noise in otherwise pure clusters.** These impurities arise from the embedding incorrectly placing faces within thresholding distance of other identities.

**Impure Cluster Detection:** Given a cluster, we estimate the average pairwise distance between faces in the cluster. The garbage clusters contain a much higher average pairwise distance than the regular identities. This is due to the similarity embedding not placing noise images closer together than a legitimate identity. Further, we found that pure clusters that contain small amounts of noise also have a significantly higher average pairwise distance, as the mean pairwise distance is not resistant to outliers. We thus use this metric (mean pairwise distance) to flag clusters as impure if they deviate by some parameter from the median metric (since the median is much more resistant) across all clusters.

To compute outliers, we use Median Absolute Deviation (MAD), which is a more robust statistic when compared to the Standard Deviation. In particular, the MAD is defined as $MAD(X) = Median(X')$ where $X'$ is the vector of absolute deviations from the median:

$$x'_i = |x_i - Median(X)|, x_i \in X, x'_i \in X'. \quad (1)$$

Thus, a cluster $c_i \in C$ with average pairwise distance $d_i \in D$ is flagged impure if:

$$\frac{|d_i - Median(D)|}{MAD(D)} > \alpha \quad (2)$$

**Inner-Cluster Purification:** Once clusters are detected as impure, it could be the case that it is a normal cluster with a small amount of noise. We purify it by searching for faces which contribute most to the pairwise distance average. I.e.,

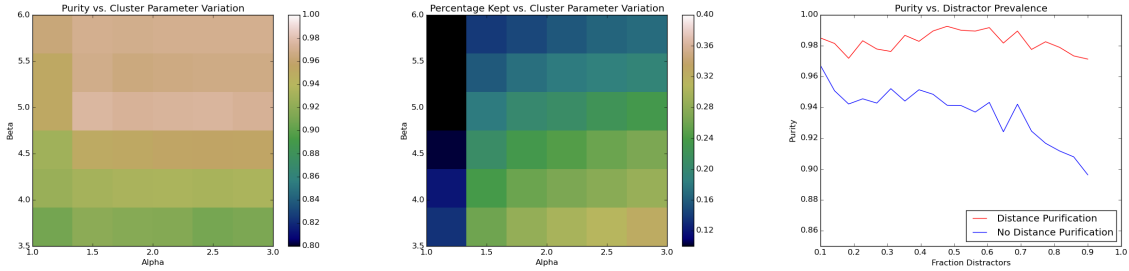

Figure 2. **Left**: Using distance optimization, cluster purity as outlier threshold $\alpha$ and distance threshold $\beta$ vary. **Center**: Using distance optimization, the fraction of photos kept as outlier threshold $\alpha$ and distance threshold $\beta$ vary. **Right**: Purity as distractor count increases with and without post-clustering optimization to detect impure clusters and performing inner-cluster purification. As expected, using purification based on distance allows maintaining cluster purity as noise increases.

We construct a distance matrix $d_k$ for a specific cluster $c_k$. The $i, j$ entry in $d_k$ is the $L_2$ between the feature vectors of faces $i$ and $j$. The matrix $d_k$, is then summed row-wise (or equivalently column-wise), such that we obtain a vector $v$ that has one score for each face measuring how much distance it contributes. We again apply MAD, and formulate a similar threshold to equation 2 to find outliers and remove them. For simplicity, we used the same threshold, $\alpha$ for both impure cluster detection and inner-cluster purification:

$$\frac{|v_i - Median(v)|}{MAD(v)} > \alpha. \quad (3)$$

These outliers are ejected from the cluster and, then once removed, we evaluate equation 2 to determine if there are still any impure clusters. If this subsequent check fails, we reject the entire cluster. Similarly to $\beta$, the parameter $\alpha$ is estimated automatically using the ground truth datasets FGNET and LFW, explained in Sec. 4.4.

### 4.4. Parameter Tuning and Validation

To validate our clustering approach and tune the $\alpha, \beta$ parameters, we constructed a series of fake clusters using labeled faces from FG-Net and LFW. We also included random samples from our collection of 1M Megaface photos for artificial noise. We sampled the parameters $(\alpha, \beta)$ on a linear scale in steps of 0.5, and for each sample pair we ran the clustering algorithm on 100 randomly generated identities. We measured 1) cluster purity and 2) fraction of faces kept. We averaged these measurements for each parameter setting. The result is a two dimensional purity surface with respect to our two clustering parameters, and an analogous surface for the fraction kept. See Figure 2 for illustration. We then selected $\beta = 5.5$ and $\alpha = 1.5$ since this pair gave the max purity (98%) and max fraction of faces kept (35%) in our experiment.

Additionally, we tested invariance to noise, i.e., testing if impure cluster detection and inner-cluster purification are robust to noise, by increasing the ratio of MegaFace distractors to legitimate identities. For this, we increased the ratio of noise to labeled images and measured the accuracy with, and without cluster distance optimization. We found that using post-clustering optimization favorably affect purity (from 90% degrading as distractors were added, to over 98% purity hardly degrading as distractors were added). The right image of Figure 2 shows the degradation of purity in the final set as distractors increase.

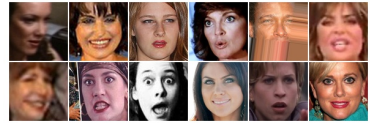

Figure 3. A set of FaceScrub impurities revealed by our outlier detection. Some due to alignment artifacts, low resolution while others are incorrect identities.

Finally, we tested how our automated purification method generalizes to other data sets. Specifically, we ran the clustering to detect outliers in FaceScrub [20] by running the optimization over each identity in FaceScrub as if it was a cluster. Even though FaceScrub is labeled, it is known to include noise. Our clustering algorithm was able to automatically detect the noisy clusters, e.g., Figure 3. These are example faces from different celebrity clusters which were found to be outliers in the embedded feature space from the rest of the cluster. 186 images were found to be outlier faces in total, representing 0.17% of the data set. No distractors were used in this experiment.

## 5. Final Dataset Statistics

In total, once clustered and optimized MF2 contains 4,753,320 faces and 672,057 identities. On average this is 7.07 photos per identity, with a minimum of 3 photos per identity, and maximum of 2469. Example identity is shown in Figure 4. We expanded the tight crop ver-sion by re-downloading the clustered faces and saving a loosely cropped version. The tightly cropped dataset requires 159GB of space, while the loosely cropped is split into 14 files each requiring 65GB for a total of 910GB. In order to gain statistics on age and gender, we ran the WIKI-IMDB [23] models for age and gender detection over the loosely cropped version of the data set. We found that females accounted for 41.1% of subjects while males accounted for 58.8%. The median gender variance within identities was 0. The average age range to be 16.1 years while the median was 12 years within identities. The distributions can be found in the supplementary material.

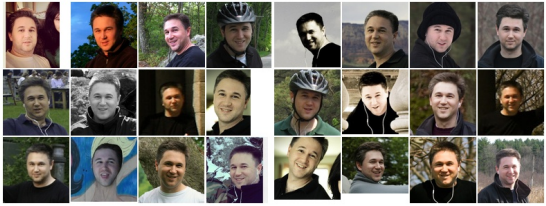

Figure 4. Example of a randomly selected identity, *78799744@N00*, in the loosely cropped version of our dataset (MF2). This cluster was not flagged as impure and therefore made it through to the final dataset. Note that the identity is found in many different lighting, expression, and camera conditions.

A trade off of this algorithm is that we must strike a balance between noise and quantity of data with the parameters. It has been noted by the VGG-Face work [21], that given the choice between a larger, more impure data set, and a smaller hand-cleaned data set, the larger can actually give better performance. A strong reason for opting to remove most faces from the initial unlabeled corpus was detection error. We found that many images were actually non-faces (see Figure 1). There were also many identities that did not appear more than once, and these would not be as useful for learning algorithms. By visual inspection of 50 randomly thrown out faces by the algorithm: 14 were non faces, 36 were not found more than twice in their respective Flickr accounts. In a complete audit of the clustering algorithm, the reason for throwing out faces are folows:

**69%** Faces which were below the $< 3$ threshold for identity
**4%** Faces which were removed from clusters as impurities
**27%** Faces which were part of clusters which were still impure even after purification

## 6. MF2 Benchmark

Participants are required to train their algorithms on MF2 dataset and test in the up to 1M distraction-probe scheme proposed by the MegaFace challenge [11]. This effectively levels the playing field between algorithms in hopes of showing which **algorithms** can perform best with the same amount of data. Specifics:

1. A participant downloads the 670K identities, and begins training their algorithm as appropriate on this data. We provide both tight crop and loose crop versions, as well as download links to the full images, face detection locations, and fiducial points.
2. The participant extracts euclidean comparable features from their trained algorithm on three different datasets: a new 1M disjoint Flickr distraction face set (containing faces which are not found in the challenge training set), FaceScrub, and FG-Net. The latter two are used as probes (since the identity is known).
3. We then evaluate each set of features as described by MegaFace Challenge: We predict several metrics (e.g. rank-1 and rank-10 identification of probe images, as well as verification) over varying levels of distractors to evaluate the algorithms' performance on the million scale.

To help ensure no test-train overlap, we produce our 1M distractor list entirely by using Flickr accounts *that did not have any samples in our final clustered set.*, and used a small amount of probe images which are either celebrities (FaceScrub) or from private study (FG-Net). If one of test identities were to appear in the Flickr data, the sparseness of our training identities (over 672K) would help mitigate the overfitting.

## 7. Baseline Training Algorithms

To provide an initial result, 4 VGG networks were used as baselines. In VGG-Face [21], training has been accomplished with class prediction per identity (softmax loss). This is equivalent to learning a function $\psi(x)$ for image $x$ which maps it to feature $l$ and is a non-linear function $\mathbb{R}^{W \times H} \rightarrow \mathbb{R}^L$. Features $\psi(x)$ are the fully connected activations just before the softmax prediction layer. $L$-dimensional feature vectors can be compared with euclidean distance or cosine similarity, but [21] showed that stronger results can be achieved by additionally learning a triplet mapping of these features, i.e., $\gamma(l)$ from feature representation $l$ to $t$ which is a mapping $\mathbb{R}^L \rightarrow \mathbb{R}^T$ where $T << L$ and is L2 normalized as a unit hypersphere. We follow the guidelines in the original paper and choose $L = 1024$ as our triplet dimension for this experiment.

Similar to experience of [8] on 100K identities, we were unable to output a prediction for all 670K identities, as the VGG network contains a softmax output layer that is fully connected. Thus, we trained the following four models:

**Model A**: Trained on random 20,000 identities (140K photos) via softmax loss. The model outputs features $\psi(x)$ are the 4096-dimensional feature activations behind the 20K softmax output layer. Model was trained for 100 epochs. In our experiments we were only able to converge training on up to 20,000 identities in the span of 48 hours. With more resources (most notably GPU memory), this approach can be scaled further.

|  | **FaceScrub** | | **FG-Net** | |
| --- | --- | --- | --- | --- |
| **Method Name** | Rank-1 | Rank-10 | Rank-1 | Rank-10 |
| GRCC | **75.772**% | **92.666**% | 21.039% | 35.781% |
| NEC | 62.122% | 78.658% | 29.294% | 43.233 % |
| Team 2009 | 58.933% | 78.724% | **38.208**% | **51.714**% |
| 3DiVi | 57.045% | 77.955% | 35.790% | 49.765% |
| VeraID | 44.191% | 61.827 % | 16.086% | 28.572% |
| TSEC | 28.716% | 43.030% | 11.566% | 22.853% |
| Baseline - Model C | 5.357% | 15.810% | 5.873% | 16.772% |
| Baseline - Model D | 3.954% | 14.326% | 2.770% | 12.694% |
| Baseline - Model A | 2.130% | 11.699% | 0.334% | 9.428% |
| Baseline - Model B | 1.846% | 11.313% | 0.189% | 9.230% |

Table 2. Rank-1 and Rank-10 identification rates for participating methods (trained on the MF2 data set). Results are reported for two probe sets: FaceScrub (celebrities) and FGNET (age invariance) with 1M distractors.

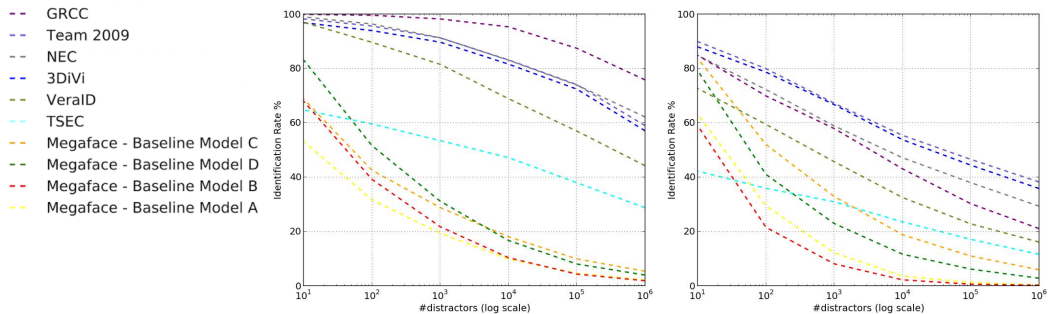

Figure 5. MF2: rank-1 **identification** rates under up to 1M distractions (varying by factors of 10) using FaceScrub (**left**) and FG-Net (**right**) as probe images. For comparison, see corresponding plots from MegaFace challenge in the sup. material.

**Model B**: Triplet tuned Model A on all 670k identities via triplet loss. An additional triplet layer is placed in front of the 4096-dimensional feature layer from a frozen Model B. The model outputs 1024-dimensional triplet features $\gamma(\psi(x))$. Model was trained for 20K iterations over the span of 24 hours (at which point the training converged).

**Model C**: Trained a *rotating softmax model* with 2,600 identities which are randomly sampled every 20th epoch. After each *rotation* the output layer is randomly re-initialized and fine-tuned (e.g. all other layers are not trained) for 5 of the 20 epochs. For the remaining 15 epochs the entire model is trained. The model outputs features $\psi(x)$ are the 4096-dimensional feature activations behind the rotating softmax output. Model was trained for 400 epochs over the course of 72 hours.

**Model D**: Apply the same triplet tuning strategy of Model B, but instead using a frozen trained Model C.

We augmented our data by randomly flipping left and right, and train on non-aligned tight crop photos (96x96), we used a learning rate of 0.001 and Adam as our optimization algorithm [13]. Triplet tuning used a learning rate of 0.25.

A 1M disjoint distraction face set is computed from the original 40M faces and used as a new MegaFace distraction set. Features are extracted by each model for this set and the two probe sets: FaceScrub and FG-Net. Rank-1 identification rates under varying distractions are shown in Figure 5. While these baselines did not account for the entire dataset they provide a useful metric for comparison. All models were trained across 4 NVIDIA Titan X GPUs.

## 8. Competition Results

Results of the groups participating in the MF2 benchmark are stated in Table 2, and Figures 5 and 6. There are a number of interesting results which we discuss in this section.

The most interesting result is that **competitors scored comparably to MegaFace and MS-Celebs-1M** where any private training sets were allowed. The highest scoring team on MF2, GRCC, obtained a rank-1 accuracy of **75.771%**, with 1M distractors in the FaceScrub probe set. By compar-ison, the highest scoring team in MegaFace, Vocord, scored **75.127%** also in FaceScrub with their private training set.

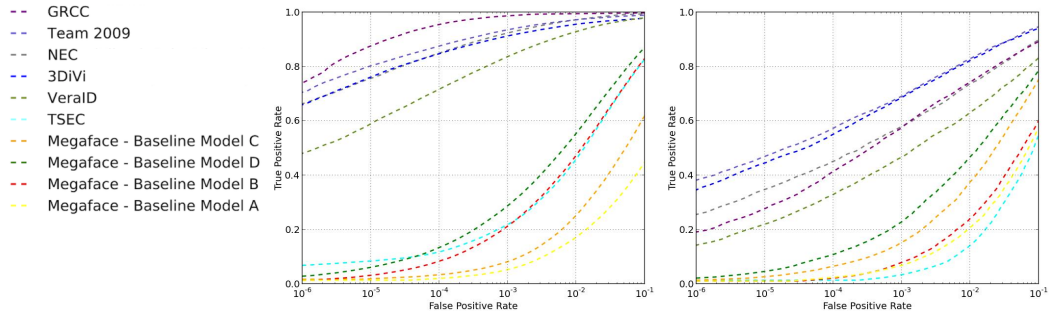

Figure 6. MF2 **verification** performance rates with 1M distractors using FaceScrub (**left**) and FG-Net (**right**) as probe images. For comparison, see corresponding plots from MegaFace challenge in the sup. material.

On the FGNET probe set (age invariant testing) algorithms perform slightly worse than when private sets are allowed. But more generally, on both challenges algorithms do not perform as well as with FaceScrub. Since on MF2 we can not claim a dataset bias (trained on celebs and tested on celebs) the issue may be deeper. It is possible that the data does not provide enough variance in age. Although the data set has people of diverse ages (shown by the distribution of ages in the Supplementary material), their individual photo collections may not span enough characteristics across ages to provide accurate recognition as obtainable by private data sets. This could be a point for future creation of training sets. Alternatively, algorithms may need to account for ages as an additional training feature.

|  | **FaceScrub** | | **FG-Net** | |
| **Trained on** | MF1 | MF2 | MF1 | MF2 |
| --- | --- | --- | --- | --- |
| 3DIVI | 33.705% | 57.045% | 15.780% | 35.790% |
| SphereFace | 75.766% | 58.933% | 47.582% | 38.208% |
| GRCCV | 77.147% | 75.772% | 24.783% | 21.039% |

Table 3. Rank-1 identification rates for methods that have both participated in MegaFace and MF2: although in the original MegaFace benchmark these algorithms may have been trained on different datasets, MF2 allows for a more fair comparison of the algorithms themselves.

Some of the groups participated in both MegaFace (MF1) and MF2, and outperformed themselves once using MF2 training data (See Table 3, which compares several groups which participated in both benchmarks). For example, the 3DIVI group increased their performance from 15% to 35% on FGNET and 35% to 57% on FaceScrub (with 1M distractors). This is a significant increase that suggests that once good training data is available to the public, algorithms can be evaluated much better. The current winner in the competition achieved around 75% in MF2 and about 74% in the MegaFace challenge [11] which suggests their algorithm is particularly good and invariant to the type of training data used.

Observing the results as a function of size of the distractor set in Figures 5 and 6 we see that, as expected, all algorithms are doing very well with 10 distractors (comparable to the LFW benchmark), and performance decreases with increase of noise (up to 1M). This is similar to MegaFace benchmark which means that MF2 is as good as other private sets used for training.

Interestingly, all three recent large-scale benchmarks report max accuracy of about 75%, and none of the state of the art methods were able to outperform that accuracy.

## 9. Summary

Advance in neural networks, have made it clear that access data is important to performance and the advancement of recognition. Small scale benchmarks and challenges are saturated. Many strong face recognition results have been achieved using techniques such as softmax loss (class prediction) as a training mechanism over a fully connected neural network layers; however, this approach scales poorly to hundreds of thousands of labels (identities). In this paper we have presented a new broad face data set featuring over 0.5M identities. We provided insights into how such a data set can be labeled and constructed and released a benchmark to level the playing field across algorithms and remove bias. Preliminary results on this competition has shown comparable results to MegaFace and MS-Celebs-1M (which allow private data sets), and important issues that need more consideration are discussed.

## Acknowledgments

The project was funded by Samsung, Google, NSF/Intel #1538613. We thank the researchers and companies that participated in the competition. Special thanks to David Porter for helping with running the large amounts of submissions and maintaining the MegaFace webpage.